\documentclass[journal]{IEEEtran}
\usepackage{amsmath,amsfonts}
\usepackage{algorithmic}
\usepackage{algorithm}
\usepackage{array}
\usepackage[caption=false,font=footnotesize,labelfont=rm,textfont=rm]{subfig}
\usepackage{textcomp}
\usepackage{stfloats}
\usepackage{url}
\usepackage{verbatim}
\usepackage{graphicx}
\usepackage{cite}
\usepackage{multirow}
\usepackage{booktabs}
\usepackage{graphicx}
\usepackage{svg}

\begin{document}

\title{Distributional Black-Box Model Inversion Attack with Multi-Agent Reinforcement Learning}

\author{Huan Bao, Kaimin Wei, Yongdong Wu, Jin Qian, Robert H. Deng,~\IEEEmembership{Fellow,~IEEE}
\thanks{H. Bao, K. Wei, Y. Wu, and J. Qian are with the College of Cyber Security, Jinan University, Guangzhou 510632, China
(e-mail: lingyan@stu2022.jnu.edu.cn, cswei@jnu.edu.cn, wuyd007@qq.com, qianjin@stu2021.jnu.edu.cn).}
\thanks{R. H. Deng are with the School of Information Systems, Singapore Management University, Singapore 178902, Singapore (e-mail: robertdeng@smu.edu.sg).}
}

\maketitle

\begin{abstract}
A Model Inversion (MI) attack based on Generative Adversarial Networks (GAN) aims to recover the private training data from complex deep learning models by searching codes in the latent space. However, they merely search a deterministic latent space such that the found latent code is usually suboptimal. In addition, the existing distributional MI schemes assume that an attacker can access the structures and parameters of the target model, which is not always viable in practice. To overcome the above shortcomings, this paper proposes a novel Distributional Black-Box Model Inversion (DBB-MI) attack by constructing the probabilistic latent space for searching the target privacy data. Specifically, DBB-MI does not need the target model parameters or specialized GAN training. Instead, it finds the latent probability distribution by combining the output of the target model with multi-agent reinforcement learning techniques. Then, it randomly chooses latent codes from the latent probability distribution for recovering the private data. As the latent probability distribution closely aligns with the target privacy data in latent space, the recovered data will leak the privacy of training samples of the target model significantly. Abundant experiments conducted on diverse datasets and networks show that the present DBB-MI has better performance than state-of-the-art in attack accuracy, K-nearest neighbor feature distance, and Peak Signal-to-Noise Ratio.
\end{abstract}


\section{Introduction}
\IEEEPARstart{A}{rtificial} Intelligence (AI) technology is rapidly advancing and widely applied in diverse domains, including facial recognition \cite{an2022killing}, autonomous driving\cite{hu2023planning}, smart homes\cite{li2023s}, drone applications\cite{zhou2022edplvo}, etc. Although AI has undoubtedly brought substantial convenience to both work and life, it is vulnerable to various attacks, such as adversarial attacks \cite{cao23,wei2022towards}, data poisoning \cite{chen2022amplifying}, \cite{tejankar2023defending} and Model Inversion (MI) attacks \cite{fredrikson2014privacy}, \cite{yin2023ginver}. Fredrikson et al. \cite{fredrikson2014privacy} demonstrated that MI attacks pose a significant risk of privacy leakage for machine learning (ML), as attackers can expose sensitive training data by only accessing the ML model itself. 

Recently, GAN-based MI attacks have emerged as an attractive way to attack complex ML models. Zhang et al. \cite{zhang2020secret} introduced the first GAN-based MI attack, shifting the focus from the algorithm-centered numerical reconstruction of sensitive data to an optimization problem through searching latent code from GAN's latent space. They utilized GAN to extract prior knowledge from publicly available datasets and searched the latent space of the GAN to recreate privacy.
GAN-based MI attacks always involve the following steps. Step 1: GAN training. It trains a GAN using the publicly available dataset that shares a similar distribution to the private dataset used by the target network. For example, if the private dataset includes facial images, the public dataset should also contain facial images. Step 2: latent code searching. It identifies the suitable latent code to generate images that could reveal private information when passed through the trained GAN.

Chen et al. \cite{chen2021knowledge} argued that previous GAN-based MI attacks are limited to one-to-one privacy recovery via the exploration of latent code. To achieve many-to-one privacy recovery, they introduced the distributional attack to reconstruct multiple privacy data instances that correspond to a single label. They proposed the Knowledge-Enriched Distributional Model Inversion (KED-MI), which initially generates pseudo-labels for a publicly available dataset using the target model. Subsequently, a GAN is trained to discriminate generated images as part of the loss function for further optimization. Finally, the trained GAN and optimized latent distribution based on the white-box setting are employed to attack the target and recreate confidential information. 
Yuan et al. \cite{yuan2023pseudo} developed the Pseudo Label-Guided MI (PLG-MI) to enhance the training method for GAN. In the GAN training, they only used photos that have a greater level of confidence in certain classes. They exclusively utilized images with higher confidence for specific classes in GAN training, enhancing the information contained within the GAN to generate images with particular labels. This will improve the GAN's ability to narrow down the search space. Although these distributional white-box MI attacks demonstrated satisfactory performance in step 1, they still have several limitations:

\begin{itemize}
    \item Requiring large-scale dataset. These attacks rely heavily on the discriminative ability of the GAN and leverage the target model to label the dataset, thus enhancing the GAN's ability to differentiate. PLG-MI, in particular, needs to examine a large amount of datasets to ensure that there is enough data for each category to train GAN. This over-reliance on prior knowledge may lead to the misuse of the target model and also increase the difficulty of dataset collection.

    \item Over-accessing the target model. KED-MI and PLG-MI assume that the attacker can freely access the parameters of the target model, enabling them to constrain the latent distribution and facilitate the identification of an appropriate latent distribution. Nevertheless, it is challenging to implement this strong assumption in real attacks. 
    
    \item Underexplored latent distribution. The latent distribution contains rich information that significantly enhances the efficacy of MI attacks. However, KED-MI and PLG-MI rely heavily on GAN, rather than latent distribution, for targeting. Thus, the under-searched latent distribution shall narrow down the attack performance of KED-MI and PLG-MI.    
\end{itemize}

To overcome the above difficulties, we propose a novel Distributional Black-Box Model Inversion (DBB-MI) attack. A GAN is trained to assign labels to datasets using a randomly chosen dataset without annotation. In the context of black-box settings, the latent distribution is optimized to effectively tackle the issue of target model over-access by utilizing Multi-Agent Reinforcement Learning (MARL) techniques. This enhances the relevance of the attacks to real-life situations. This paper presents the primary contributions as follows:

\begin{itemize}
    \item We propose the Distributional Black-Box Model Inversion (DBB-MI) Attack, which is the first exploration of a distributional MI attack in black-box settings.

    \item In black-box settings, DBB-MI leverages the Multi-Agent Reinforcement Learning (MARL) algorithm to thoroughly explore the appropriate latent distributions for specific categories, extracting latent privacy features in GAN. 
    
    \item Extensive experiments have demonstrated the superior attack performance of DBB-MI compared with state-of-the-art black-box MI attacks. 
    For example, its highest success rate has experienced a notable boost of 33.6\% on CelebA. Additionally, it also achieves a 100\% attack success rate on MNIST.
    \end{itemize}

The rest of this paper is arranged as follows. Section \ref{1026:related} introduces some related work. Section \ref{1026:approach} gives the challenges of searching for latent distribution under the black-box setting and provides a detailed description of DBB-MI. Section \ref{1026:exper} presents and analyzes experimental results. Section \ref{min:conclusion} concludes this work.

\section{Related Work}
\label{1026:related}
This section introduces the background knowledge of Multi-Agent Reinforcement Learning (MARL) and Model Inversion (MI) Attacks.
\begin{figure}[t]
    \centering
    \includegraphics[width=0.75\linewidth]{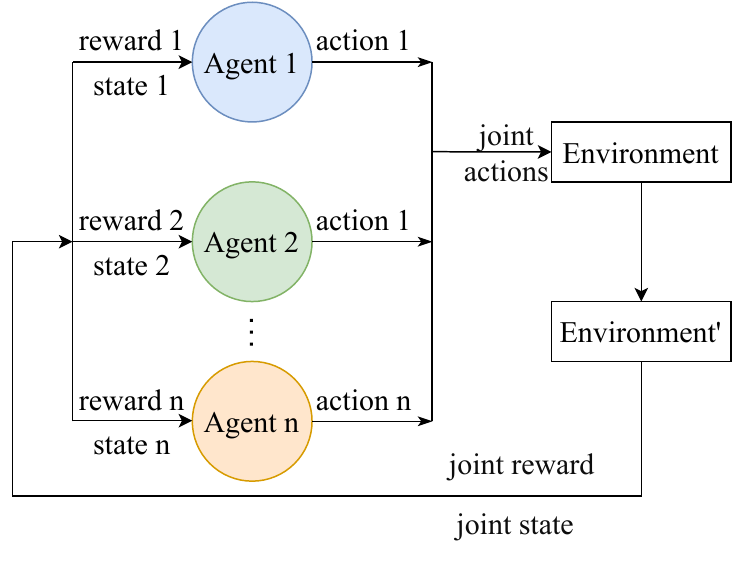}
     \caption{The overview of multi-agent reinforcement learning.}
     \label{fig:marl}
\end{figure}

\subsection{Multi-Agent Reinforcement Learning}
Single-agent reinforcement learning employs the Markov decision process model, whereas multi-agent reinforcement learning (MARL) incorporates stochastic games. The joint actions formed by multiple agents have a significant impact on the transition and updating of the environmental state. Additionally, these actions play a crucial role in determining the reward feedback received by the agents, as depicted in Figure \ref{fig:marl}. The agents can be categorized into three groups based on their relationships: totally cooperative, fully competitive, and semi-cooperative semi-competitive.

In the totally cooperative MARL \cite{gupta2017cooperative, 10.5555/3237383.3238080}, all agents are dedicated to jointly achieving a shared objective by maximizing the overall reward through collaboration, without considering their individual rewards. For example, each agent possesses its own local value function in the Value-Decomposition Networks (VDN) \cite{10.5555/3237383.3238080} algorithm. After each agent makes a decision, the local value is calculated and then aggregated to obtain the global value to achieve globally optimal choices.

In fully competitive MARL \cite{zhu2020online, tampuu2017multiagent}, all agents see each other as competitors, and each agent only focuses on maximizing its utility, disregarding the impact of other agents. One illustrative instance is the Independent Q-Learning (IQL) algorithm \cite{tampuu2017multiagent}, wherein agents independently engage in Q-learning. Although it may produce satisfactory outcomes in certain contexts, it often exhibits instability and difficulties in achieving convergence due to the influence of other actors on the surrounding environment. Therefore, it is typically only suitable for relatively simple scenarios.

In semi-cooperative semi-competitive MARL \cite{hu2003nash,lowe2017multi}, agents can obtain greater benefits through collaboration, while simultaneously experiencing potential gains or losses due to a certain level of competition. The Multi-Agent Deep Deterministic Policy Gradient (MADDPG) \cite{lowe2017multi} algorithm enables each agent not only to learn its policy knowledge but also to observe the behaviors of other agents during the learning process to improve their strategy further. These algorithms are more suitable for handling complex multi-agent tasks because they allow agents to simultaneously assess cooperative and competitive connections.

Through the above analysis, it is easy to get that different MARL approaches have different advantages and adapted environments. When utilizing MARL, it is crucial to choose the appropriate MARL according to the surroundings and the particular activity in order to accomplish the goals more effectively.

\subsection{Model Inversion Attacks}

According to attack strategies, existing MI attacks can be divided into direct reconstruction-based and GAN-based.

Early direct reconstruction-based MI attacks predominantly concentrated on white-box MI attacks, in which an attacker can access all data related to the model, including architecture, parameters, and others. Fredrikson et al. \cite{fredrikson2014privacy} developed the first MI attack, targeting the information regression model by inputting specific features. However, its effectiveness diminishes when the feature space increases. Later, Fredrikson et al. \cite{fredrikson2015model} achieved the inversion of a face dataset with a larger feature space by minimizing the confidence loss. Although white-box MI attacks can disclose the privacy of the target model, they assume that an attacker can access anything about the target model, which is the opposite of reality. 
Therefore, some researchers focused on black-box MI attacks, in which an adversary only possesses the outputs or labels of the model rather than all the information. Yang et al. \cite{yang2019neural} assumed that the adversary has access to a vast database that far exceeds the training data of the target network. They employed this data as auxiliary information for an attack. Additionally, Salem et al. \cite{salem2020updates} attacked the newly added training data by comparing different outputs on the same data before and after the target model was updated. Zhang et al. \cite{zhang2023analysis} enhanced face reconstruction accuracy by fully exploiting predicted vectors. Nevertheless, direct reconstruction-based MI attacks can only recover grayscale images with substantial information loss on simple networks. 

To attack deep networks, Zhang et al. \cite{zhang2020secret} proposed the GAN-based MI attack, seeking potential private data within the latent space of GAN to gain the target network's privacy.
Currently, GAN-based MI attacks can be further divided into two subcategories: optimizing latent code and optimizing latent distribution. MI attacks based on optimizing latent code are essentially black-box attacks. An et al. \cite{an2022mirror} utilized genetic algorithms to implement GAN-based MI attacks, reconstructing high-fidelity private face images within deep networks. Han et al. \cite{han2023reinforcement} achieved impressive MI attacks by utilizing reinforcement learning algorithms to search for latent codes within the latent space of GANs. Zhu et al. \cite{zhu2022label} utilized the error rate of the target model to explore decision boundaries, reconstructing representative samples. Kahla et al. \cite{kahla2022label} proposed the Boundary-Repelling Model Inversion Attack (BERP-MI), which only uses GAN to generate images with a target label and extracts the latent space of images to gather sufficient data for estimating the gradient direction. Due to the limited private information in the latent code, GAN-based MI attacks that rely on optimizing latent code have challenges in accurately recovering results.


To obtain more private information, some researchers focus on MI attacks based on optimizing latent distribution, which are actually white-box attacks. Chen et al. \cite{chen2021knowledge} proposed the Knowledge-Enriched Distributional Model Inversion Attack (KED-MI), which leverages the target network to generate pseudo-labels for GAN training to boost the attack rates significantly. Yuan et al. \cite{yuan2023pseudo} further refined the training methodology of GAN by selecting more representative data from public datasets, thereby enhancing the capability of GAN and narrowing the search space within the latent space. Moreover, meticulously chosen datasets contain a greater abundance of privacy features, consequently further improving the attack performance.



Since these MI attacks based on optimizing latent distribution are all white-box attacks, their requirements on the dataset and the assumption of using the target model to label the dataset are still too unrealistic. In this paper, we attempt to develop a distributional black-box MI attack that does not require a super attacker and the elaborate training of GANs. 

\section{The Proposed Approach: DBB-MI}
\label{1026:approach}
In this section, we introduce the challenges of searching for latent distribution under the black-box setting and provide a detailed description of DBB-MI.

\subsection{Problem Formulation}

\begin{figure*}
    \centering
    \includegraphics[width=0.6\linewidth]{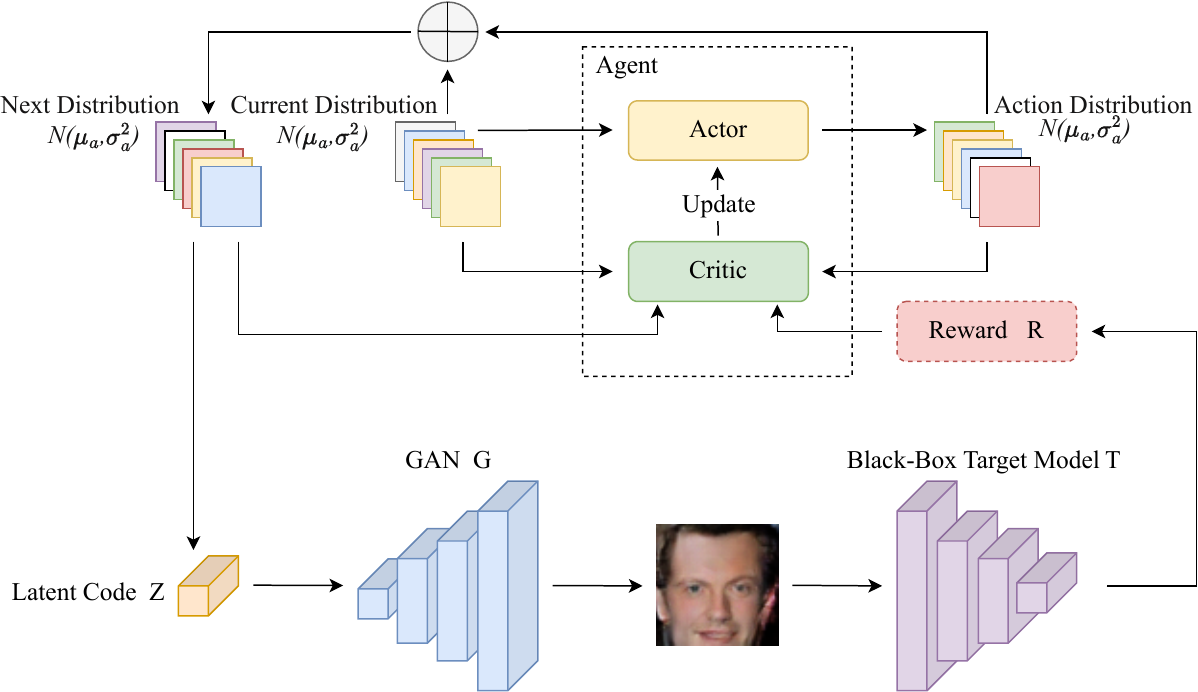}
    \caption{The overview of DBB-MI. To search for the target latent space distribution, two agents are employed to optimize the target distribution's $\mu$ and $\sigma$, respectively.}
    \label{fig:overview}
\end{figure*}

\begin{figure}[t]
    \centering
    \includegraphics[width=0.7\linewidth]{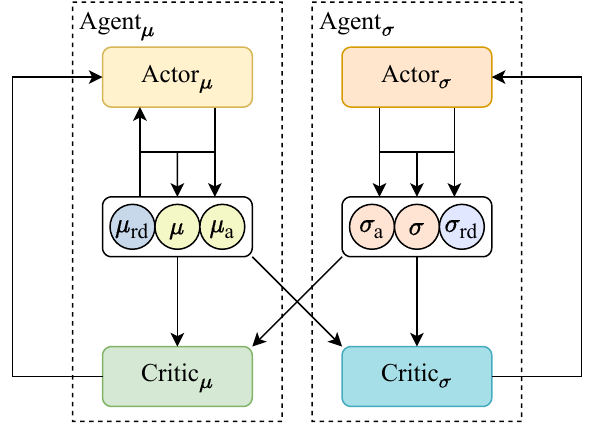}
     \caption{The overall collaboration and competition between two agents. The actor network makes decisions by observing the environmental state, while the critic network feeds feedback to the actor network according to its global observations.
     }
     \label{fig:maddpg}
\end{figure}

\subsubsection{Attack model}
This work primarily examines black-box MI attacks, where the attacker can neither access the private data $D_{priv}$ of the black-box target network $T$ nor obtain any knowledge about the model's structure, hyper-parameters, etc.
The black-box target model $T$ is trained to recognize $k$ different identities using a private dataset $D_{priv}$. The model $T$ produces a probability distribution according to the following way:
\begin{equation}
    T(x) \rightarrow [0,1]^k
\end{equation}
where $[0,1]^k$ represents the probabilities of $x$ being classified into each of the $k$ identities.

We can only obtain the corresponding $T(x)$ by inputting an image $x$ into the target model for discrimination in the black-box model without obtaining any intermediate parameters from any of the models.
The attacker aims to obtain sensitive data $x$ associated with a specific label $y$ from the target network. We chose the face recognition classifier model $T$ as the attack target to make our attack more realistic. This model identifies individuals' identities in images and assigns the corresponding labels. Thus, the private facial images of any specific identity are constructed by utilizing the soft and hard labels provided by the black-box model.

It is essential to satisfy the following conditions to expose the privacy of the target $y$ by reconstructing data $x$: 1) The probability of the target label $y$ in the prediction probability distribution of model $T$ on input $x$ must be maximized, i.e., $argmax_y (T(x)_y) = y$. 2) The confidence level of the label $y$ should be as large as possible, i.e., maximizing $T(x)_y$.

\subsubsection{Stochastic Game for Latent Distribution Search}

In GANs, the variation of latent code in the latent space is continuous. Hence, searching for latent code can be regarded as a Markov decision process (MDP). However, searching for latent distribution cannot be considered an MDP. Firstly, latent distribution has a more complex state space than latent code, and the parameters constituting the distribution, such as mean $\mu$ and variance $\sigma$, entail more uncertainty and interaction. Therefore, searching for latent distribution should be viewed as a stochastic game.

When dealing with stochastic game problems, multi-agent reinforcement learning is often a good choice as it can effectively address interactions and competitions. Therefore, we select two agents to optimize the mean $\mu$ and variance $\sigma$ constituting the latent distribution, respectively. In the context of distributional MI attacks, these two agents aim to optimize the appropriate latent distribution for selecting appropriate latent code to reconstruct more privacy-preserving images. Even so, we cannot classify this task as entirely cooperative. A certain degree of competition exists between $\mu$ and $\sigma$. Maintaining this competitive dynamic enables them to enhance their performance while continuously striving for global optimality. This competitive relationship fosters flexibility in the optimization process, ultimately leading to improved MI attack performance. Therefore, we choose the Multi-Agent Deep Deterministic Policy Gradient (MADDPG) \cite{lowe2017multi} as the MARL agent for searching appropriate latent distribution.

\subsubsection{Overview} 
DBB-MI consists of three steps. Firstly, we train a GAN, where the GAN $G$ is initially trained on the public dataset $D_{pub}$. It's important to note that the public dataset $D_{pub}$ does not overlap with the private dataset $D_{priv}$ used to train the target black-box model $T$.
We neither need to select the images carefully nor to use the target model $T$ for additional labelling of images in $D_{pub}$.
Next, we optimize the initial random latent distribution to approximate the real latent distribution. Finally, we sample the latent code from the optimized high-dimensional latent distribution and input it into the GAN $G$ to reconstruct private data. The overall structure of DBB-MI is exhibited in Fig.\ref{fig:overview}.

\subsection{MADDPG for Searching Latent Space Distribution}
\subsubsection{MADDPG}
The MADDPG \cite{lowe2017multi} agent has two fundamental components: the actor and critic networks, as depicted in Fig.\ref{fig:maddpg}. The actor network determines the actions to be executed by the agent based on the information observed by the agent as well as the current state of the system. The critic network is responsible for judging the value of actions and providing feedback (i.e., a reward signal) to the actor. This process enables the actor network to update its policies. Through iterations of the above operations, the agent gradually learns to optimize parameters $\mu$ and $\sigma$ to minimize the discrepancy between the optimized and real latent distribution.

\subsubsection{Action}
The actor network aggregates all the data the agent has observed, including the environment's current state, other agents' observations, and other pertinent information. Based on this information, the actor network makes decisions regarding the agent's actions to optimize the latent distribution parameters $\mu$ and $\sigma$. We independently model the latent distribution of each dimension of the latent code and then sample each dimension independently from these latent distributions. Fig.\ref{fig:overview} shows that $\mu$ and $\sigma$ are sampled from the standard normal distribution to form the initial random distribution $N(\mu,\sigma)$. In addition, actions $action_\mu$ and $action_\sigma$ are selected by $Actor_\mu$ and $Actor_\sigma$ according to the initial parameters.

\begin{equation}
    \mu_a=Actor(N(\mu,\sigma^2))
    \label{eq:action}
  \end{equation}
  
The following technique is employed to update actions:
\begin{equation}
    \mu_{t+1}=\alpha\mu_t+(1-\alpha)\mu_a
    \label{eq:endac}
\end{equation}

Previous research \cite{han2023reinforcement,wang2021variational} has demonstrated that exploring latent space impacts the diversity and accuracy of reconstructed images. So, we introduce a parameter $\alpha$ to balance accuracy and diversity. A small $\alpha$ value is employed in the early stages of searching to encourage the agent to optimize the distribution, broadening the search scope. As the training process advances, the latent distribution optimized by agents eventually approaches the real latent space distribution. The $\alpha$ is gradually increased to mitigate the diversity of the generated images. The optimization procedure enables agents to refine the latent distribution further to generate a latent distribution closely related to natural images.

\subsubsection{Reward}
The critic network evaluates the value and utility of the actions by measuring their consistency with the target task. When optimizing the latent distribution, the critic network provides feedback or rewards to motivate agents to take actions to enable it toward the real latent space. Actions that steer the latent distribution toward the real latent space will be rewarded more significantly; otherwise, only a lower or no reward will be given.

As evidenced in Fig.\ref{fig:maddpg}, the critic network considers the effect of actions on the state of the environment. It is updated by incorporating environmental feedback, actual rewards, and estimated action rewards. These help to improve the estimation of rewards by the critic networks. The agents move closer to the true latent distribution by optimizing actor and critic networks. When the current distribution is closer to the real latent distribution, the images generated from that distribution will have higher confidence in the target network $T$. Therefore, the reward can be calculated as follows:
\begin{equation}
    r_{t+1}=log[T_l(G(z_{t+1}~N(\mu_{t+1},{\sigma_{t+1}}^2)))]
    \label{eq:rewardto}
\end{equation}
\begin{equation}
    r_{a}=log[T_l(G(z_{a}~N(\mu_{a},{\sigma_{a}}^2)))]
    \label{eq:rewardac}
\end{equation}
where $r_{t+1}$ denotes the score of images generated from the latent space of the new distribution after performing actions, and $r_{a}$ is the reward that agents receive after performing actions.

In optimizing the latent distribution, it is necessary to compute rewards individually for each action based on its effectiveness and impact on the target. The reward calculation way helps improve the precision of dynamic adjustments in the optimization process, as formulated in the following:

\begin{equation}
    r_{\mu}=log[T_l(G(z_{\mu}~N(\mu_{t+1},\sigma^2_{t})))]
    \label{eq:rewardmu}
\end{equation}
\begin{equation}
    r_{\sigma}=log[T_l(G(z_{\sigma}~N(\mu_{t},\sigma^2_{t+1})))]
    \label{eq:rewardsig}
\end{equation}

We also introduce the penalty factor $r_c$ to penalize instances where the generated image is irrelevant to the target category.
This could help agents perform actions to improve the quality of the generated image while reducing interference from non-target categories. 
The penalty term can assist agents in optimizing the distribution so that the reconstructed images are closely related to the target category, yielding superior-quality images. The penalty factor $r_c$ is defined below.

\begin{equation}
    r_c=max(\varepsilon,-p_{l})
    \label{eq:penalty}
\end{equation}

\begin{figure}[t]
    \centering
    \includegraphics[width=0.65\linewidth]{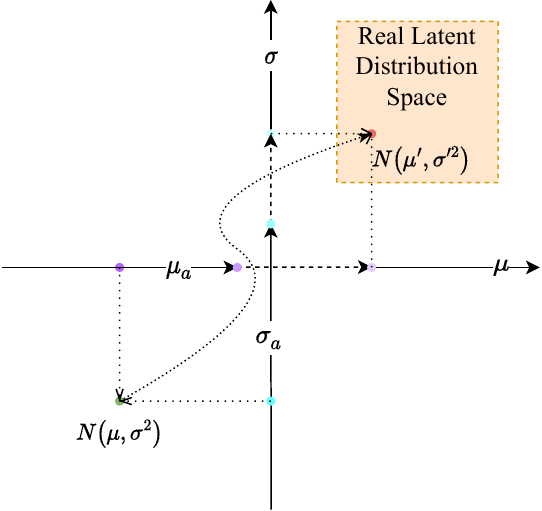}
     \caption{The overall steps of one-dimensional latent distribution search. From random latent distribution to real latent distribution.}
     \label{fig:laten}
  \end{figure}

The threshold $\varepsilon$ is introduced to prevent agents from obtaining additional rewards.
When the negative log probability ($-p_l$) of the target label of the generated image exceeds the specific threshold, an additional penalty is imposed. This penalty factor ensures that the generated image is more relevant to the target category and that images are distinguishable enough to avoid confusion with non-target categories.

To sum up, we can calculate rewards for agent$_\mu$ and agent$_\sigma$ as follows:
\begin{equation}
    R_\mu=w_1r_{t+1}+w_2r_a+w_3r_\mu+w_4r_c
\end{equation}
\begin{equation}
    R_\sigma=w_1r_{t+1}+w_2r_a+w_3r_\sigma+w_4r_c
\end{equation}
where w$_n$ represents the weight of r$_n$.

\subsubsection{Distribution optimization}
As shown in Fig.\ref{fig:laten}, the search is conducted on the high-dimensional latent distribution using MADDPG. Specifically, 
the $\mu = \{\mu_1, \mu_2, ..., \mu_n\}$ and $\sigma = \{\sigma_1, \sigma_2, ..., \sigma_n\}$ are sampled from an $n$-dimensional normal distribution. After that, they are paired together to form the initial $n$-dimensional high-dimensional latent distribution as follows:
\begin{equation}
    L = N(\mu, \sigma^2); \mu \sim N(0, I), \sigma \sim N(0, I)
\end{equation}

The MADDPG algorithm, involving $Agent_\mu$ and $Agent_\sigma$, optimizes the initial high-dimensional latent distribution. $Agent_\mu$ and $Agent_\sigma$ select actions based on the current initial distribution, and these actions are formulated below.
\begin{equation}
    \mu_a = \{\mu_{a1}, \mu_{a2}, ..., \mu_{an}\}
\end{equation}
\begin{equation}
    \sigma_a^2 = \{\sigma_{a1}^2, \sigma_{a2}^2, \ldots, \sigma_{an}^2\}
\end{equation}

The chosen actions facilitate the latent distribution $L$ towards the real latent distribution, resulting in a highly optimized latent distribution $L' = N(\mu', {\sigma'}^2)$. The latent code $z'$ is ultimately extracted from $L'$.
\begin{equation}
    z' = \{z'_1, z'_2, ..., z'_n\}
\end{equation}
where $z'_i$ obeys the distribution $N(\mu'_i, {\sigma'_i}^2)$. 

The above optimization method of high-dimensional latent distribution can avoid the limitation caused by directly searching the latent distribution space.
Through MARL methods, an effective search of the latent distribution can be achieved with only limited model outputs, without the need for any additional information about the model.
Consequently, in a black-box setting, it broadens the search range in the latent distribution space, thereby enhancing the ability to identify the real latent space of the target and ultimately improving the accuracy of the reconstructed sensitive data.

\subsection{MADDPG for Agent Training}
The MADDPG trains two agents, enabling them to cooperate and compete in a predefined image generation environment. Agents select specific actions to optimize the initial latent distribution toward the real latent distribution.
The key lies in training the agents, which specifically involves the following steps:
\begin{itemize}
    \item \textbf{Step 1:} After observing the randomly
    constructed initial distribution $N(\mu,\sigma^2)$,
    the agents select corresponding actions to execute;
    \item  \textbf{Step 2:} The rewards for the decisions made
    by the agent are computed and stored in the
    replay buffer $B$ along with the agent's observation information;
    \item \textbf{Step 3:} 
    When there are enough experiences in the replay buffer $B$, a batch is sampled to update the Agent,
    and the updated Agent is returned.
\end{itemize}
The private data hidden in the target network is reconstructed. Details about agent training are provided in Algorithm \ref{alg:maddpg}.
    
\begin{algorithm}[H]
    \caption{MADDPG for Agent Training}
    \label{alg:maddpg}
    \begin{algorithmic}[1]
        \REQUIRE Target Model: $T$, Target Label: $l$, GANs: $G$
        \ENSURE Trained agents: $agent_\mu$, $agent_\sigma$
        \STATE Initialize new $agent_\mu$, $agent_\sigma$, replay buffer $B$
        \FOR{$round=1$ to $max\_rounds$}
            \STATE Initialize $n$ dims vector $\mu_t$, $\sigma_t$
            \FORALL{$\omega$ in $\mu,\sigma$}
                \STATE $\omega_a \leftarrow Actor_\omega(N(\mu_t,\sigma_t^2))$
                \STATE $\omega_{r+1} \leftarrow \alpha\omega_t+(1-\alpha)\omega_a$
                \STATE $r_{t+1} \leftarrow log[T_l(G(z_{t+1}\sim N(\mu_{t+1},\sigma_{t+1}^2)))]$
                \STATE $r_a \leftarrow log[T_l(G(z_{a}\sim N(\mu_{a},\sigma_{a}^2)))]$
                \STATE \verb|//| Obtaining $z_\omega$ under different $\omega$
                \STATE $ // z_{\mu}\sim N(\mu_{t+1},\sigma_{t}^2),z_{\sigma}\sim N(\mu_{t},\sigma_{t+1}^2)$.
                \STATE $r_\omega \leftarrow log[T_l(G(z_{\omega}))]$
                \STATE $r_c \leftarrow max(\varepsilon,-p_l)$
                \STATE $R_\omega \leftarrow w_1r_{t+1}+w_2r_a+w_3r_\omega+w_4r_c$
            \ENDFOR
            \STATE Add ($\mu_t$,$\mu_a$,$\mu_{t+1}$,$R_\mu$,$\sigma_t$,$\sigma_a$,$\sigma_{t+1}$,$R_\sigma$) to $B$.
            \IF{$len(B) > max\_len$}
                \STATE Sample a random mini-batch from $B$.
                \STATE Calculate the actor loss and critic loss.
                \STATE Update the actor and critic networks.
            \ENDIF 
        \ENDFOR
        \RETURN $agent_\mu$,$agent_\sigma$
    \end{algorithmic}
\end{algorithm}

\section{Experiment}
\label{1026:exper}

In this section, we primarily analyze the attack performance of DBB-MI on different datasets and target networks. Additionally, we analyze the distributional attack and investigate some factors that may affect the attack performance.

\subsection{Experimental setting}
\subsubsection{Dataset}Four distinct face datasets that represent a variety of situations are used to evaluate the effectiveness and breadth of DBB-MI.
Additionally, we conducted experiments on the MNIST dataset to assess the applicability of our approach to other types of datasets.

\begin{itemize}
    \item CelebFaces Attributes Dataset (CelebA) \cite{liu2015deep}. It contains 202,599 photos of 10,177 different celebrities.

    \item FaceScrub \cite{ng2014data}. It includes 106,863 images of 530 individuals with an even gender distribution.

    \item Pubfig83 \cite{pinto2011scaling}. It consists of 13,600 images of 83 individuals. These images were taken in regulated real-world environments with significant variations in lighting, expressions, and other attributes.

    \item Flickr-Faces-HQ (FFHQ) \cite{karras2019style}. It comprises 70,000 high-quality face images with significant age, expression, and ethnicity variations.

    \item MNIST. It encompasses 70,000 handwritten digits (0 through 9), having different structures and features from facial datasets.
\end{itemize}

CelebA, FaceScrub, Pubfig83, and MNIST are divided into two parts: public and private datasets. The public dataset is utilized to train GAN, while the private dataset is employed to train the target classification model. It should be emphasized that there is no overlap of the same identities or images between public and private datasets. Thus, it can be assumed that the trained GAN does not directly contain any original private information. Additionally, since public and private datasets in the same dataset have similar statistical properties, the FFHQ is used as an independent extra dataset to evaluate the performance of MI attacks under various distribution conditions. It allows for a comprehensive assessment of MI attacks' robustness and generalization capabilities.

\subsubsection{Target Models}
Like previous studies \cite{chen2021knowledge,zhang2020secret,kahla2022label,han2023reinforcement}, we utilize three popular face recognition networks for evaluation; namely FaceNet64 \cite{cheng2017know}, ResNet-152 \cite{he2016deep}, and VGG16 \cite{2015Very}. These networks are employed to assess the impact of MI attacks on models with different architectures. The generalization and robustness of DBB-MI are better evaluated using varied face recognition models.

\subsubsection{Baselines}
We select some representative state-of-the-art white-box and black-box MI attacks as baselines for comparison. Specifically, we choose the Generative Model Inversion (GMI) attacks \cite{zhang2020secret} and Knowledge-Enriched Distributional Model Inversion (KED-MI) \cite{chen2021knowledge} attacks as white-box MI attacks. GMI is the first MI attack for deep networks, while KED-MI is a distributional MI attack. Meanwhile, we employ the Reinforcement Learning-based Black-box Model Inversion (RLB-MI) attacks \cite{han2023reinforcement} and Model Inversion for deep learning Network (MIRROR) \cite{an2022mirror}, representing the advanced black-box MI attack. We also select the Boundary-Repelling Model Inversion (BERP-MI) attacks \cite{kahla2022label}, the only and most advanced label-only MI attack. These black-box MI attacks represent the current state-of-the-art (SOTA) in GAN-based MI attacks that directly search for latent code.

All models undergo identical dataset training, and the same evaluation models assess all experimental results to ensure fair comparisons. GMI, RLB-MI, MIRROR, and BERP-MI utilize the same GAN as DBB-MI. In addition, the GAN is trained for KED-MI   using the specified requirements and an identical dataset to that of DBB-MI. It allows a fair and objective comparison between KED-MI and DBB-MI.

\begin{figure*}
    \centering
    \includegraphics[width=0.8\linewidth]{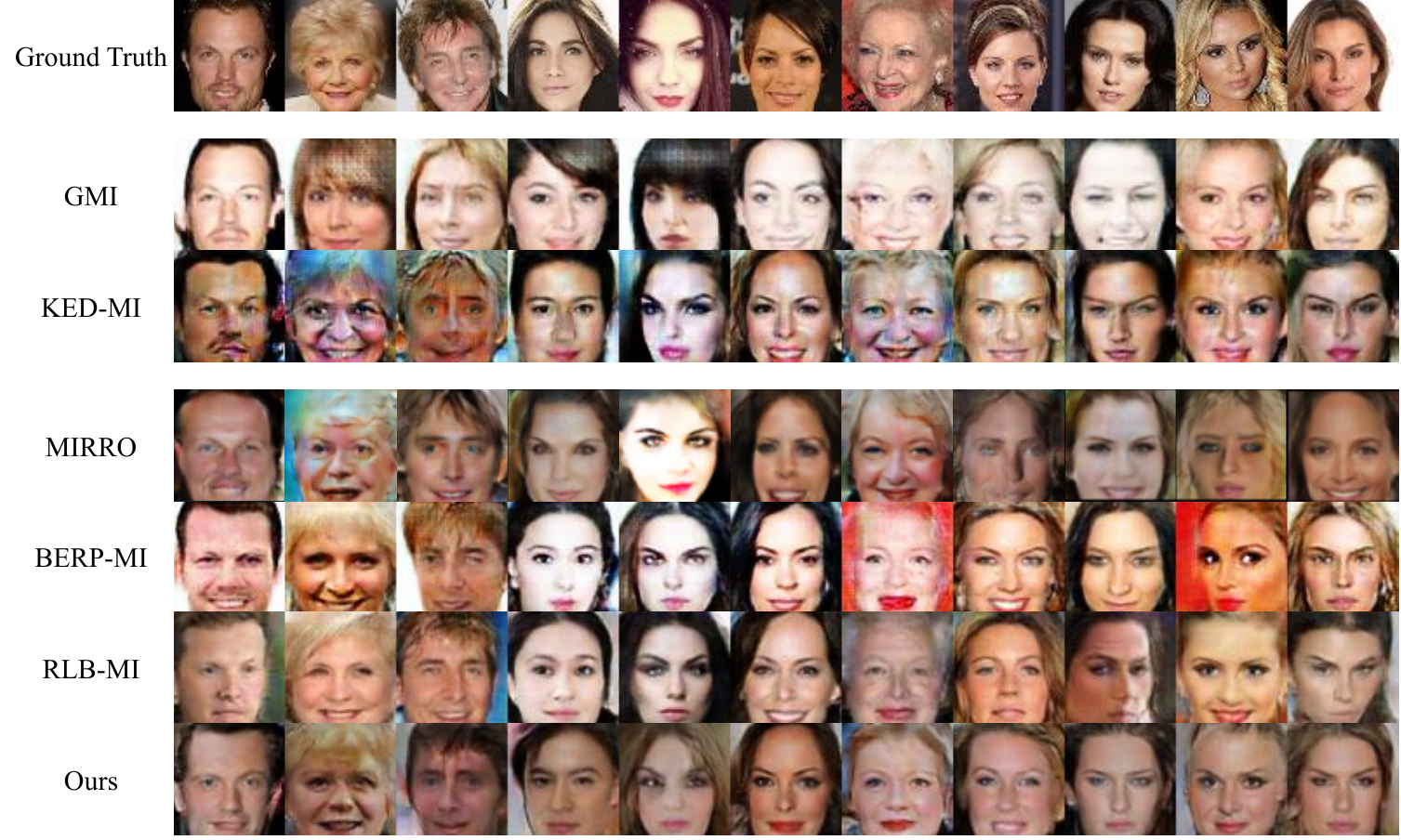}
    \caption{The images reconstructed by different MI attacks under CelebA and VGG16. The top row displays the real images, the middle two rows show the images reconstructed by the white-box MI baselines, and the bottom four rows exhibit the images reconstructed by the black-box MI baselines and our method DBB-MI.}
    \label{fig:compare}
\end{figure*}

\subsubsection{Implementation details}
The same hyperparameters are used to train the GAN and target network like previous studies \cite{chen2021knowledge,zhang2020secret,kahla2022label,han2023reinforcement}. For MADDPG, some important parameters are set as follows:
\begin{itemize}
    \item learning rate: 1e-3
    \item discount factor: 0.99
    \item target network update rate: 5\texttimes$10^{-3}$
    \item experience replay buffer size: 1\texttimes$10^{6}$
    \item batch size: 256
    \item training episodes: 4\texttimes$10^{4}$
\end{itemize}

\subsubsection{Evaluation metrics}
Like prior work \cite{zhang2020secret}, the effectiveness of MI attacks is evaluated using quantitative criteria, including attack accuracy (ACC) and K-nearest neighbor feature distance (KNN Dist). Furthermore, the Peak Signal-to-Noise Ratio (PSNR) is utilized to assess the resemblance between reconstructed and original images.

\textit{Attack Accuracy:}This metric measures the probability of successfully reconstructing private data through an attack. The key with \cite{chen2021knowledge,han2023reinforcement} difference is that we only consider the attack successful when both the target model and the additional discriminative model agree that the generated image belongs to the target class. This approach enhances the accuracy of attack success rate assessment, ensuring that the generated images deceive the target model and exhibit high-quality facial features, reducing cases where noisy images are mistakenly identified as the target class.

\textit{KNN Dist:} This metric measures the similarity of features between the generated reconstructed images and real private images. To calculate the KNN Dist, features are first extracted from the fully connected layer of the evaluation classifier for both the generated reconstructed images and the real private images. Then, their similarity in the feature space is assessed by calculating the L2 distance between these two sets of features.

\textit{PSNR:} This metric measures the difference between two images. It evaluates the quality and similarity between the reconstructed and real private images by calculating their PSNR value. A higher PSNR value indicates less difference between the two images, implying a higher similarity between the reconstructed and real private images.

\begin{table}[]
\caption{The experimental results of MI attacks on different target models trained CelebA. The symbols $\uparrow$ and $\downarrow$ denote that higher and lower scores give better attack performance, respectively. The best-performing attack metrics are marked in bold.}
\scriptsize
\centering
    \begin{tabular*}{0.95\linewidth}{cl|cl|clclclcl}
        \toprule
    \multicolumn{2}{c|}{Model}                      & \multicolumn{2}{c|}{Type}  & \multicolumn{2}{c|}{Method}       & \multicolumn{2}{c}{ACC{$\uparrow$}}            & \multicolumn{2}{c}{PSNR{$\uparrow$}}           & \multicolumn{2}{c}{KNN Dist{$\downarrow$}}         \\
    \midrule
    \multicolumn{2}{c|}{\multirow{6}{*}{VGG16}}    & \multicolumn{2}{c|}{\multirow{2}{*}{White-box}}  & \multicolumn{2}{c|}{GMI}          & \multicolumn{2}{c}{0.194}          & \multicolumn{2}{c}{12.3}           & \multicolumn{2}{c}{1521.05}          \\
    \multicolumn{2}{c|}{}               &\multicolumn{2}{c|}{}             & \multicolumn{2}{c|}{KED-MI}        & \multicolumn{2}{c}{0.684}          & \multicolumn{2}{c}{14.59}          & \multicolumn{2}{c}{1258.65}          \\
    \cmidrule{3-12}
    \multicolumn{2}{c|}{}              & \multicolumn{2}{c|}{\multirow{4}{*}{Black-box}}              & \multicolumn{2}{c|}{MIRROR}        & \multicolumn{2}{c}{0.452}          & \multicolumn{2}{c}{14.21}          & \multicolumn{2}{c}{1358.20}          \\
    \multicolumn{2}{c|}{}                        &\multicolumn{2}{c|}{}    & \multicolumn{2}{c|}{BERP-MI}      & \multicolumn{2}{c}{0.562}          & \multicolumn{2}{c}{13.36}          & \multicolumn{2}{c}{1872.48}          \\
    \multicolumn{2}{c|}{}                       &\multicolumn{2}{c|}{}     & \multicolumn{2}{c|}{RLB-MI}      & \multicolumn{2}{c}{0.642}          & \multicolumn{2}{c}{15.80}           & \multicolumn{2}{c}{1262.34}          \\
    \multicolumn{2}{c|}{}                       &\multicolumn{2}{c|}{}     & \multicolumn{2}{c|}{\textbf{DBB-MI}} & \multicolumn{2}{c}{\textbf{0.858}} & \multicolumn{2}{c}{\textbf{20.66}} & \multicolumn{2}{c}{\textbf{1180.63}} \\
    \midrule
    \multicolumn{2}{c|}{\multirow{6}{*}{FaceNet64}} & \multicolumn{2}{c|}{\multirow{2}{*}{White-box}}  & \multicolumn{2}{c|}{GMI}          & \multicolumn{2}{c}{0.298}          & \multicolumn{2}{c}{15.78}          & \multicolumn{2}{c}{1584.24}          \\
    \multicolumn{2}{c|}{}                    &\multicolumn{2}{c|}{}        & \multicolumn{2}{c|}{KED-MI}        & \multicolumn{2}{c}{0.766}          & \multicolumn{2}{c}{16.35}          & \multicolumn{2}{c}{1411.56}          \\
    \cmidrule{3-12}
    \multicolumn{2}{c|}{}              & \multicolumn{2}{c|}{\multirow{4}{*}{Black-box}}               & \multicolumn{2}{c|}{MIRROR}        & \multicolumn{2}{c}{0.528}          & \multicolumn{2}{c}{15.09}          & \multicolumn{2}{c}{1308.40}          \\
    \multicolumn{2}{c|}{}                   &\multicolumn{2}{c|}{}        & \multicolumn{2}{c|}{BERP-MI}      & \multicolumn{2}{c}{0.734}          & \multicolumn{2}{c}{13.69}          & \multicolumn{2}{c}{1685.29}          \\
    \multicolumn{2}{c|}{}                  &\multicolumn{2}{c|}{}          & \multicolumn{2}{c|}{RLB-MI}      & \multicolumn{2}{c}{0.804}          & \multicolumn{2}{c}{16.27}          & \multicolumn{2}{c}{1354.86}          \\
    \multicolumn{2}{c|}{}                &\multicolumn{2}{c|}{}            & \multicolumn{2}{c|}{\textbf{DBB-MI}} & \multicolumn{2}{c}{\textbf{0.916}} & \multicolumn{2}{c}{\textbf{18.06}} & \multicolumn{2}{c}{\textbf{1091.15}} \\
    \midrule
    \multicolumn{2}{c|}{\multirow{6}{*}{ResNet-152}} & \multicolumn{2}{c|}{\multirow{2}{*}{White-box}}  & \multicolumn{2}{c|}{GMI}          & \multicolumn{2}{c}{0.340}           & \multicolumn{2}{c}{15.36}          & \multicolumn{2}{c}{1752.15}          \\
    \multicolumn{2}{c|}{}                 &\multicolumn{2}{c|}{}           & \multicolumn{2}{c|}{KED-MI}        & \multicolumn{2}{c}{0.826}          & \multicolumn{2}{c}{16.52}          & \multicolumn{2}{c}{1130.05}          \\
    \cmidrule{3-12}
    \multicolumn{2}{c|}{}                   & \multicolumn{2}{c|}{\multirow{4}{*}{Black-box}}          & \multicolumn{2}{c|}{MIRROR}        & \multicolumn{2}{c}{0.640}          & \multicolumn{2}{c}{15.91}          & \multicolumn{2}{c}{1254.90}          \\
    \multicolumn{2}{c|}{}               &\multicolumn{2}{c|}{}             & \multicolumn{2}{c|}{BERP-MI}      & \multicolumn{2}{c}{0.754}          & \multicolumn{2}{c}{13.17}          & \multicolumn{2}{c}{1745.73}          \\
    \multicolumn{2}{c|}{}                    &\multicolumn{2}{c|}{}        & \multicolumn{2}{c|}{RLB-MI}      & \multicolumn{2}{c}{0.812}          & \multicolumn{2}{c}{15.23}          & \multicolumn{2}{c}{1308.69}          \\
    \multicolumn{2}{c|}{}                      &\multicolumn{2}{c|}{}      & \multicolumn{2}{c|}{\textbf{DBB-MI}} & \multicolumn{2}{c}{\textbf{0.898}} & \multicolumn{2}{c}{\textbf{17.37}} & \multicolumn{2}{c}{\textbf{1063.38}} \\
    \bottomrule
    \end{tabular*}
    \label{tab:CelebA} 
\end{table}

\subsection{Comparison with state-of-the-art MI attacks}

\subsubsection{Performance evaluation on different target models}
Table \ref{tab:CelebA} shows the experimental results of our method and baselines under different target models. As seen in Table \ref{tab:CelebA}, DBB-MI exhibits a notable superiority compared to state-of-the-art white-box and black-box MI attacks regarding ACC, KNN Dist, and PSNR. Using the target model VGG16 as an example, the ACC of DBB-MI improves 25.4\% over KED-MI and 33.6\% over RLB-MI. This is because DBB-MI fully explores the latent space by optimizing the latent distribution and obtaining more private data about the target. The experimental results in Table \ref{tab:CelebA} demonstrate that DBB-MI is more effective in targeting different target models and poses more severe privacy leakage risks. This indicates that our method outperforms white-box distributional attacks and achieves SOTA black-box attack performance.

In addition, we also compare the images reconstructed by DBB-MI with those rebuilt by baselines. As displayed in Fig.\ref{fig:compare}, the images recovered by DBB-MI are closer to the original ones compared to those recovered by baselines; they have similar details and colors as the original. It is attributed to the efficiency of DBB-MI in searching the latent space, allowing it to capture more private information. The above experimental results prove that DBB-MI outperforms the state-of-the-art white-box and black-box MI attacks for various target models in terms of multiple performance evaluation metrics and visualization.

\begin{table}[]
\caption{The experimental results of MI attacks on different datasets.}
\scriptsize
\centering
    \begin{tabular*}{0.95\linewidth}{cc|cc|cc|cccccc}
      \toprule
    \multicolumn{2}{c|}{Dataset}         & \multicolumn{2}{c|}{Type}              & \multicolumn{2}{c|}{Method}  & \multicolumn{2}{c}{ACC{$\uparrow$}}   & \multicolumn{2}{c}{PSNR\textbf{$\uparrow$}}  & \multicolumn{2}{c}{KNN Dist{$\downarrow$}} \\ 
    \midrule
    \multicolumn{2}{c|}{\multirow{6}{*}{CelebA}}   & \multicolumn{2}{c|}{\multirow{2}{*}{White-box}}   & \multicolumn{2}{c|}{GMI}     & \multicolumn{2}{c}{0.298} & \multicolumn{2}{c}{15.78} & \multicolumn{2}{c}{1584.24}  \\
    \multicolumn{2}{c|}{}                  &\multicolumn{2}{c|}{}          & \multicolumn{2}{c|}{KED-MI}   & \multicolumn{2}{c}{0.766} & \multicolumn{2}{c}{16.35} & \multicolumn{2}{c}{1411.56}  \\
    \cmidrule{3-12}
    \multicolumn{2}{c|}{}                & \multicolumn{2}{c|}{\multirow{4}{*}{Black-box}}             & \multicolumn{2}{c|}{MIRROR}   & \multicolumn{2}{c}{0.528} & \multicolumn{2}{c}{15.09} & \multicolumn{2}{c}{1308.40}  \\
    \multicolumn{2}{c|}{}               &\multicolumn{2}{c|}{}             & \multicolumn{2}{c|}{BERP-MI} & \multicolumn{2}{c}{0.734} & \multicolumn{2}{c}{13.69} & \multicolumn{2}{c}{1685.29}  \\
    \multicolumn{2}{c|}{}                &\multicolumn{2}{c|}{}            & \multicolumn{2}{c|}{RLB-MI} & \multicolumn{2}{c}{0.804} & \multicolumn{2}{c}{16.27} & \multicolumn{2}{c}{1354.86}  \\
    \multicolumn{2}{c|}{}                &\multicolumn{2}{c|}{}            & \multicolumn{2}{c|}{\textbf{DBB-MI}}     & \multicolumn{2}{c}{\textbf{0.916}} & \multicolumn{2}{c}{\textbf{18.06}} & \multicolumn{2}{c}{\textbf{1091.15}}  \\ 
    \midrule
    \multicolumn{2}{c|}{\multirow{6}{*}{FaceScurb}}   & \multicolumn{2}{c|}{\multirow{2}{*}{White-box}} & \multicolumn{2}{c|}{GMI}     & \multicolumn{2}{c}{0.080}  & \multicolumn{2}{c}{17.05} & \multicolumn{2}{c}{2729.06}   \\
    \multicolumn{2}{c|}{}                &\multicolumn{2}{c|}{}            & \multicolumn{2}{c|}{KED-MI}   & \multicolumn{2}{c}{0.355} & \multicolumn{2}{c}{20.31} & \multicolumn{2}{c}{2682.69}   \\
    \cmidrule{3-12}
    \multicolumn{2}{c|}{}                 & \multicolumn{2}{c|}{\multirow{4}{*}{Black-box}}            & \multicolumn{2}{c|}{MIRROR}   & \multicolumn{2}{c}{0.325} & \multicolumn{2}{c}{18.86} & \multicolumn{2}{c}{2710.11}  \\
    \multicolumn{2}{c|}{}                  &\multicolumn{2}{c|}{}          & \multicolumn{2}{c|}{BERP-MI} & \multicolumn{2}{c}{0.305} & \multicolumn{2}{c}{20.81} & \multicolumn{2}{c}{2684.91}   \\
    \multicolumn{2}{c|}{}                &\multicolumn{2}{c|}{}            & \multicolumn{2}{c|}{RLB-MI} & \multicolumn{2}{c}{\textbf{0.420}}  & \multicolumn{2}{c}{19.23} & \multicolumn{2}{c}{2693.85}   \\
    \multicolumn{2}{c|}{}                &\multicolumn{2}{c|}{}            & \multicolumn{2}{c|}{\textbf{DBB-MI}}     & \multicolumn{2}{c}{0.375} & \multicolumn{2}{c}{\textbf{23.03}} & \multicolumn{2}{c}{\textbf{2661.82}}   \\ 
    \midrule
    \multicolumn{2}{c|}{\multirow{6}{*}{Pubfig83}}  & \multicolumn{2}{c|}{\multirow{2}{*}{White-box}}  & \multicolumn{2}{c|}{GMI}     & \multicolumn{2}{c}{0.100} & \multicolumn{2}{c}{10.26} & \multicolumn{2}{c}{2580.71}   \\
    \multicolumn{2}{c|}{}               &\multicolumn{2}{c|}{}             & \multicolumn{2}{c|}{KED-MI}   & \multicolumn{2}{c}{0.380}  & \multicolumn{2}{c}{15.25} & \multicolumn{2}{c}{2363.12}   \\
    \cmidrule{3-12}
    \multicolumn{2}{c|}{}                 & \multicolumn{2}{c|}{\multirow{4}{*}{Black-box}}            & \multicolumn{2}{c|}{MIRROR}   & \multicolumn{2}{c}{0.300} & \multicolumn{2}{c}{13.25} & \multicolumn{2}{c}{2410.62}  \\
    \multicolumn{2}{c|}{}              &\multicolumn{2}{c|}{}              & \multicolumn{2}{c|}{BERP-MI} & \multicolumn{2}{c}{0.400}  & \multicolumn{2}{c}{13.10}  & \multicolumn{2}{c}{2492.99}   \\
    \multicolumn{2}{c|}{}               &\multicolumn{2}{c|}{}             & \multicolumn{2}{c|}{RLB-MI} & \multicolumn{2}{c}{0.400}  & \multicolumn{2}{c}{16.37} & \multicolumn{2}{c}{2349.84}   \\
    \multicolumn{2}{c|}{}              &\multicolumn{2}{c|}{}              & \multicolumn{2}{c|}{\textbf{DBB-MI}}     & \multicolumn{2}{c}{\textbf{0.560}}  & \multicolumn{2}{c}{\textbf{17.04}} & \multicolumn{2}{c}{\textbf{2342.47}}   \\ 
    \bottomrule
  \end{tabular*}
      \label{tab:FaceNet64}  
\end{table}

\subsubsection{Performance evaluation on different dataset}
Table \ref{tab:FaceNet64} presents the experimental results of DBB-MI and baselines under different datasets. It can be observed from Table \ref{tab:FaceNet64} that DBB-MI beats baselines in all performance evaluation metrics. Taking CelebA as a case study, DBB-MI's ACC is 19.5\% and 13.9\% higher than KED-MI and RLB-MI, respectively.

The experimental outcomes obtained by all MI attacks on CelebA are superior to those on FaceScrub and Pubfig83. This is because CelebA contains more identity categories and data information, giving the trained model stronger classification capabilities and, therefore, more vulnerability to attacks. Additionally, DBB-MI outperforms all baselines on all datasets except FaceScrub. On FaceScrub, the ACC of DBB-MI is slightly lower than that of RLB-MI, but the PSNR and KNN Dist of DBB-MI are better than those of RLB-MI. Thus, DBB-MI outperforms RLB-MI in most performance evaluation metrics.

\begin{table}[]
\caption{The experimental results of MI attacks using the GAN trained on FFHQ.}
\scriptsize
\centering
    \begin{tabular*}{0.95\linewidth}{cc|cc|cc|cccccc}
        \toprule
    \multicolumn{2}{c|}{Dataset}      & \multicolumn{2}{c|}{Type}           & \multicolumn{2}{c|}{Method}  & \multicolumn{2}{c}{ACC{$\uparrow$}}            & \multicolumn{2}{c}{PSNR{$\uparrow$}}           & \multicolumn{2}{c}{KNN Dist{$\downarrow$}}        \\
    \midrule
    \multicolumn{2}{c|}{\multirow{6}{*}{CelebA}} & \multicolumn{2}{c|}{\multirow{2}{*}{White-box}} & \multicolumn{2}{c|}{GMI}     & \multicolumn{2}{c}{0.114}          & \multicolumn{2}{c}{15.35}          & \multicolumn{2}{c}{1431.45}         \\
    \multicolumn{2}{c|}{}                &\multicolumn{2}{c|}{}         & \multicolumn{2}{c|}{KED-MI}   & \multicolumn{2}{c}{0.408}          & \multicolumn{2}{c}{\textbf{17.52}} & \multicolumn{2}{c}{1035.31}         \\
    \cmidrule{3-12}
    \multicolumn{2}{c|}{}                & \multicolumn{2}{c|}{\multirow{4}{*}{Black-box}}          & \multicolumn{2}{c|}{MIRROR}   & \multicolumn{2}{c}{0.286}          & \multicolumn{2}{c}{16.66} & \multicolumn{2}{c}{1242.91}         \\
    \multicolumn{2}{c|}{}               &\multicolumn{2}{c|}{}          & \multicolumn{2}{c|}{BERP-MI} & \multicolumn{2}{c}{0.398}          & \multicolumn{2}{c}{15.01}          & \multicolumn{2}{c}{1331.25}         \\
    \multicolumn{2}{c|}{}             &\multicolumn{2}{c|}{}            & \multicolumn{2}{c|}{RLB-MI} & \multicolumn{2}{c}{0.402}          & \multicolumn{2}{c}{16.76}          & \multicolumn{2}{c}{\textbf{925.61}} \\
    \multicolumn{2}{c|}{}             &\multicolumn{2}{c|}{}            & \multicolumn{2}{c|}{\textbf{DBB-MI}}     & \multicolumn{2}{c}{\textbf{0.532}} & \multicolumn{2}{c}{16.04}          & \multicolumn{2}{c}{1002.51}         \\
    \midrule
    \multicolumn{2}{c|}{\multirow{6}{*}{FaceScrub}}  & \multicolumn{2}{c|}{\multirow{2}{*}{White-box}}  & \multicolumn{2}{c|}{GMI}     & \multicolumn{2}{c}{0.150}          & \multicolumn{2}{c}{11.33}          & \multicolumn{2}{c}{2892.42}          \\
    \multicolumn{2}{c|}{}              &\multicolumn{2}{c|}{}           & \multicolumn{2}{c|}{KED-MI}   & \multicolumn{2}{c}{0.315}          & \multicolumn{2}{c}{15.28}          & \multicolumn{2}{c}{2856.36}          \\
    \cmidrule{3-12}
    \multicolumn{2}{c|}{}             & \multicolumn{2}{c|}{\multirow{4}{*}{Black-box}}             & \multicolumn{2}{c|}{MIRROR}   & \multicolumn{2}{c}{0.285}          & \multicolumn{2}{c}{15.52} & \multicolumn{2}{c}{2878.36}         \\
    \multicolumn{2}{c|}{}              &\multicolumn{2}{c|}{}           & \multicolumn{2}{c|}{BERP-MI} & \multicolumn{2}{c}{0.295}          & \multicolumn{2}{c}{15.97}          & \multicolumn{2}{c}{2859.94}          \\
    \multicolumn{2}{c|}{}             &\multicolumn{2}{c|}{}            & \multicolumn{2}{c|}{RLB-MI} & \multicolumn{2}{c}{0.355}          & \multicolumn{2}{c}{16.56}          & \multicolumn{2}{c}{2866.43}          \\
    \multicolumn{2}{c|}{}              &\multicolumn{2}{c|}{}           & \multicolumn{2}{c|}{\textbf{DBB-MI}}     & \multicolumn{2}{c}{\textbf{0.490}} & \multicolumn{2}{c}{\textbf{17.04}} & \multicolumn{2}{c}{\textbf{2817.85}} \\
    \midrule
    \multicolumn{2}{c|}{\multirow{6}{*}{PubFig83}} & \multicolumn{2}{c|}{\multirow{2}{*}{White-box}}  & \multicolumn{2}{c|}{GMI}     & \multicolumn{2}{c}{0.080}          & \multicolumn{2}{c}{11.16}          & \multicolumn{2}{c}{2684.99}          \\
    \multicolumn{2}{c|}{}              &\multicolumn{2}{c|}{}           & \multicolumn{2}{c|}{KED-MI}   & \multicolumn{2}{c}{0.340}           & \multicolumn{2}{c}{14.59}          & \multicolumn{2}{c}{2415.76}          \\
    \cmidrule{3-12}
    \multicolumn{2}{c|}{}            & \multicolumn{2}{c|}{\multirow{4}{*}{Black-box}}              & \multicolumn{2}{c|}{MIRROR}   & \multicolumn{2}{c}{0.300}          & \multicolumn{2}{c}{14.27} & \multicolumn{2}{c}{2459.85}         \\
    \multicolumn{2}{c|}{}            &\multicolumn{2}{c|}{}             & \multicolumn{2}{c|}{BERP-MI} & \multicolumn{2}{c}{0.380}           & \multicolumn{2}{c}{15.36}          & \multicolumn{2}{c}{2391.67}          \\
    \multicolumn{2}{c|}{}              &\multicolumn{2}{c|}{}           & \multicolumn{2}{c|}{RLB-MI} & \multicolumn{2}{c}{0.360}            & \multicolumn{2}{c}{16.05}          & \multicolumn{2}{c}{2402.5}           \\
    \multicolumn{2}{c|}{}            &\multicolumn{2}{c|}{}             & \multicolumn{2}{c|}{\textbf{DBB-MI}}     & \multicolumn{2}{c}{\textbf{0.700}}   & \multicolumn{2}{c}{\textbf{16.55}} & \multicolumn{2}{c}{\textbf{2387.56}}\\
        \bottomrule
    \end{tabular*}
    \label{tab:FFHQ}    
    \end{table}

\begin{figure}[t]
        \centering
        \includegraphics[width=0.9\linewidth]{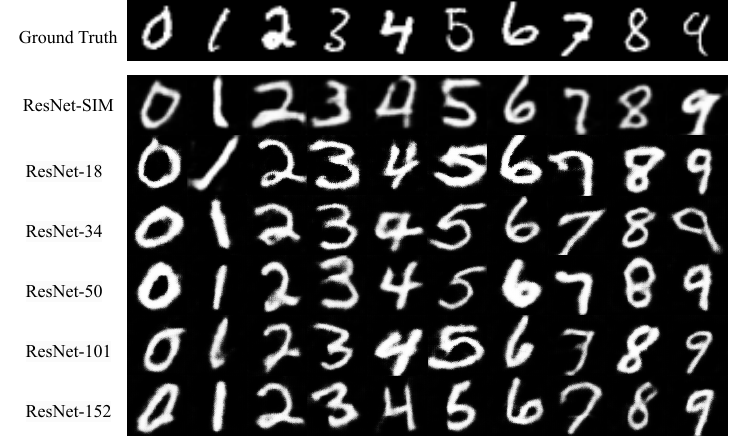}
         \caption{
         The reconstruction results of MNIST.}
         \label{fig:mnist}
      \end{figure}
      
\begin{figure}[!t]
    \centering
    \subfloat[Initial Distribution]{\includegraphics[width=0.435\linewidth]{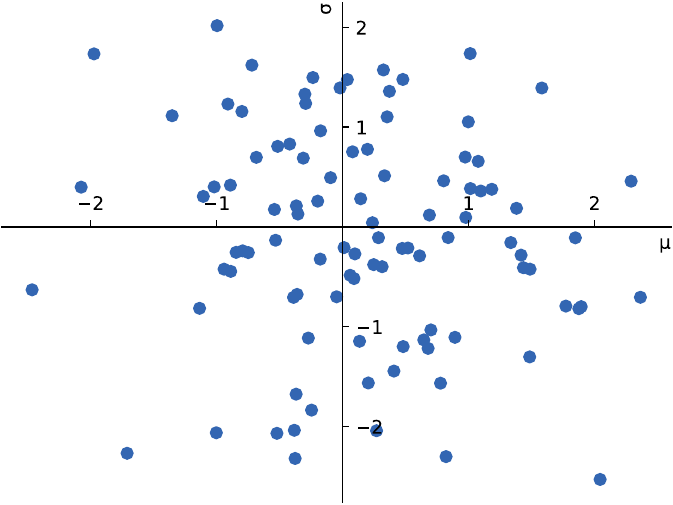}%
    \label{fig:star}}
    \hfil
    \subfloat[Final Distribution]{\includegraphics[width=0.48\linewidth]{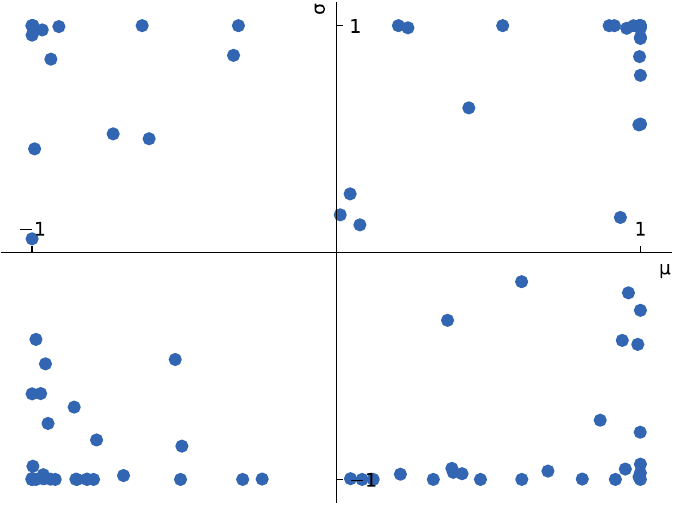}%
    \label{fig:end}}
     \caption{The initial and final latent space distribution. (a) depicts the initial latent distribution, and (b) shows the resulting latent distribution after optimization.}
     \label{fig:cp}
  \end{figure}

 \begin{figure}[t]
        \centering
        \includegraphics[width=0.85\linewidth]{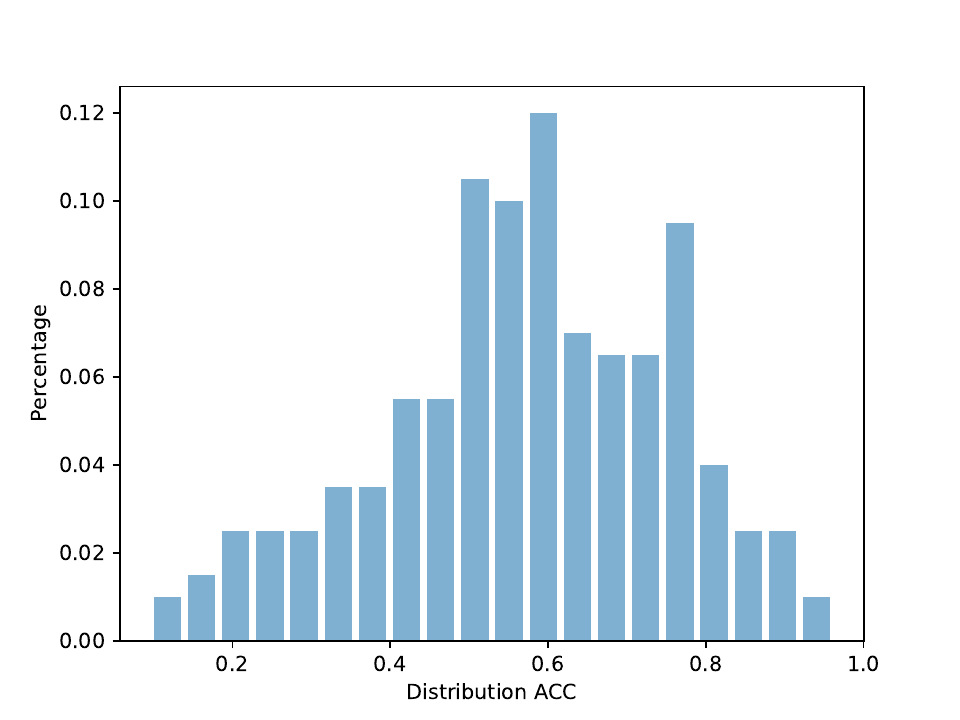}
         \caption{Distribution of different accuracy levels. The results obtained from randomly attacking 200 target labels.}
         \label{fig:disacc}
      \end{figure}

\subsubsection{Performance evaluation on cross-dataset}
In previous experiments, we utilized a dataset with similar statistical properties and feature distributions as the dataset used to train the target model to train GAN. However, obtaining a dataset with similar distributions to the target dataset in real-world settings is challenging. Therefore, it is imperative to train GAN using an extra dataset.

We train GAN on the extra dataset, FFHQ. Meanwhile, we employ the FaceNet64 trained under CelebA, FaceScrub, and Pubfig83 as the target models and utilize the FaceNet trained on these datasets as the evaluation models.
Table \ref{tab:FFHQ} presents the experimental results of DBB-MI and baselines across datasets. It is easy to see that DBB-MI has superior performance compared to baselines across most datasets. The reason for this can be linked to the comprehensive exploration of the latent space in DBB-MI, which has resulted in the acquisition of more private data about the target. In addition, DBB-MI has the highest ACC on CelebA, while its PSNR and KNN Dist are slightly worse than the best. This implies that DBB-MI still has room for performance improvement.

In summary, DBB-MI exhibits superior performance across diverse target models, datasets, and cross-dataset scenarios. It outperforms state-of-the-art white-box and black-box MI attacks regarding ACC, KNN Dist, PSNR, and visualization. These findings confirm the effectiveness of the latent distribution exploration in DBB-MI, which is vital for improving model security and privacy protection.

\subsubsection{Performance evaluation on MNIST}
To demonstrate that our approach is practical for face datasets, we also evaluated it on the MNIST dataset. For the MNIST dataset, we attacked models of different depths, including ResNet-18, ResNet-34, ResNet-50, ResNet-101, ResNet-152, and ResNet-SIM. ResNet-SIM consists of two convolutional layers, one max-pooling layer, two residual blocks, and one fully connected layer, while the remaining network structures conform to \cite{he2016deep}. We achieved a 100\% attack success rate across different networks, and features of different digits could be reconstructed. Specific reconstruction results are shown in Fig
\ref{fig:mnist}.

\subsection{Analysis of distributional attacks}

\subsubsection{Analysis of the changes of the latent distribution}
Fig. \ref{fig:star} depicts the latent distribution before attacking, in which each point represents a specific dimension of the latent distribution. During an attack, the latent distribution of each dimension is optimized by the MADDPG to explore the latent space effectively. Fig.\ref{fig:end} illustrates the latent distribution after attacking, in which each dimension of the latent distribution displays a distinct pattern. This implies that distinct privacy features are evident in each dimension of latent distribution. Moreover, it also proves the rationality of DBB-MI, i.e., optimizing each dimension of the latent distribution independently to enable the latent distribution to approach the true one.

\subsubsection{Evaluating Distribution Accuracy}
We conducted 500 random samples from the optimized latent distribution to obtain reconstructed samples. We calculated the proportion of recreated samples that matched the target labels to measure the accuracy of the optimized distribution. The accuracy of the optimized distribution was assessed by calculating the fraction of rebuilt samples that corresponded to the target labels, which is referred to as distributional accuracy. 68\% of all latent distributions have an accuracy exceeding 0.5, indicating that more than half of the randomly sampled samples can be deemed successful reconstructions. As displayed in Fig.\ref{fig:disacc}, the specific results indicate that the optimized latent distribution demonstrates strong performance in effectively exploring the target label information.

 \begin{figure}[t]
        \centering
        \includegraphics[width=0.85\linewidth]{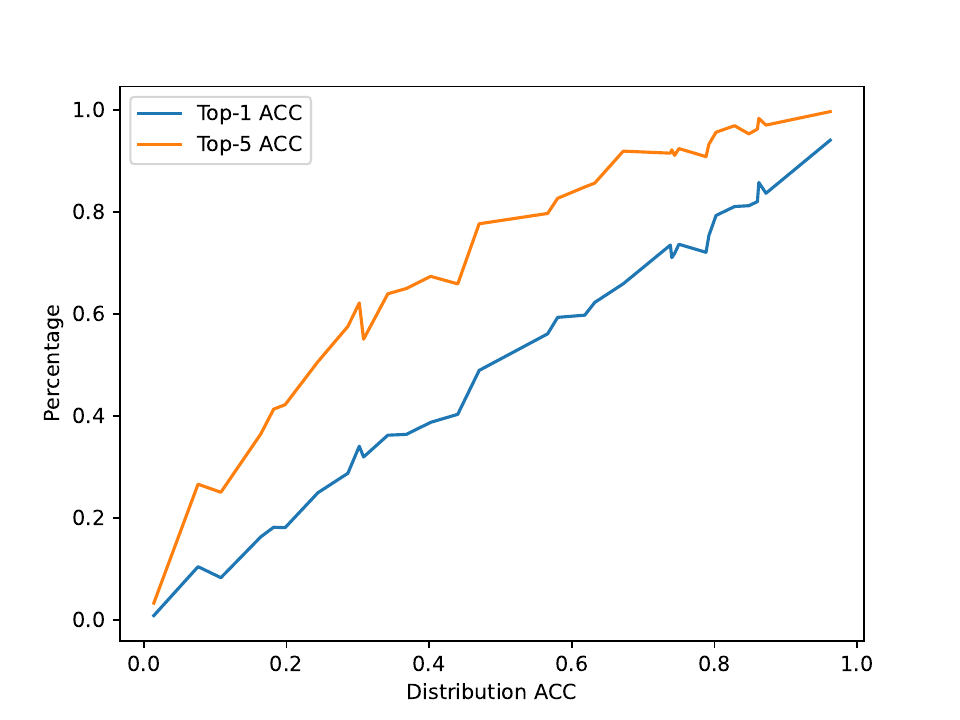}
         \caption{Actual accuracy levels.}
         \label{fig:actcc}
      \end{figure}
\subsubsection{Evaluating actual attack performance under different distributional accuracy levels}
To depict the real attack accuracy, we performed 10,000 random samples for all optimized latent distributions. These samples were then tested for top-1 accuracy, which indicates successful reconstruction, as well as top-5 accuracy, which signifies the target label ranking among the top 5 out of all 1,000 classes. When the latent distribution accuracy exceeds 24\%, our latent distribution's top-5 accuracy exceeds 50\%. The observed top-1 accuracy closely aligns with the accuracy of the tested distribution, suggesting that the optimized distribution demonstrates consistent performance in all reconstruction tasks. More details as shown in Fig.\ref{fig:actcc}.

 \begin{figure}[t]
        \centering
        \includegraphics[width=0.85\linewidth]{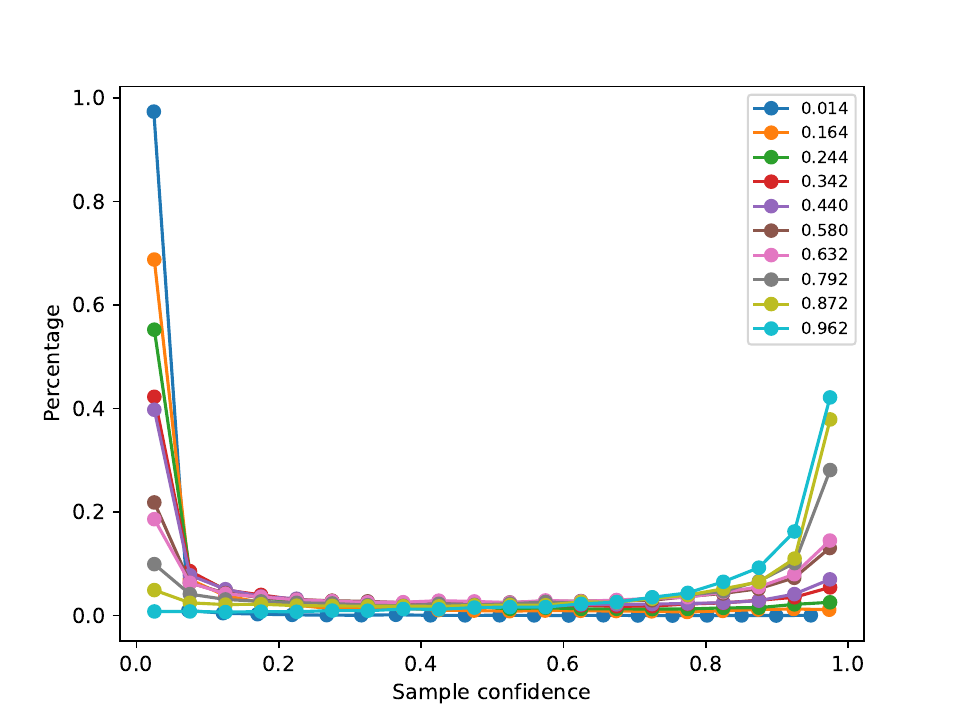}
         \caption{The relationship between distributional accuracy and reconstruction accuracy.}
         \label{fig:acttest}
      \end{figure}
\subsubsection{Evaluating sample reconstruction confidence under varying distributional accuracies}
For testing, we selected samples from 10,000 random samplings of optimized latent distributions with varying accuracies. We evaluated the reconstructed samples using the target model to determine their corresponding label confidences, as depicted in Fig.\ref{fig:acttest}. When the distributional accuracy reached 0.632, it was observed that 50.59\% of the rebuilt samples achieved a confidence level of 0.5, while 36.10\% exhibited a confidence level of 0.75. However, upon attaining a distributional accuracy of 0.962, it was seen that 90.27\% of the samples demonstrated a confidence level of 0.5, 78.58\% exhibited a confidence level of 0.75, and 58.40\% demonstrated a confidence level beyond 0.9. Despite the restricted distributional accuracy of 0.580, a significant proportion of the samples, specifically 47.51\%, surpassed a confidence level of 0.5. Although there is an improvement in the accuracy of reconstructed samples as the distributional accuracy improves, it is important to note that certain samples still exhibit very low confidence levels. This may be attributed to the inadequate training of the GAN model employed in our study.

\subsection{Analysis of factors affecting attack performance}

    \begin{figure}[t]
        \centering
        \includegraphics[width=0.85\linewidth]{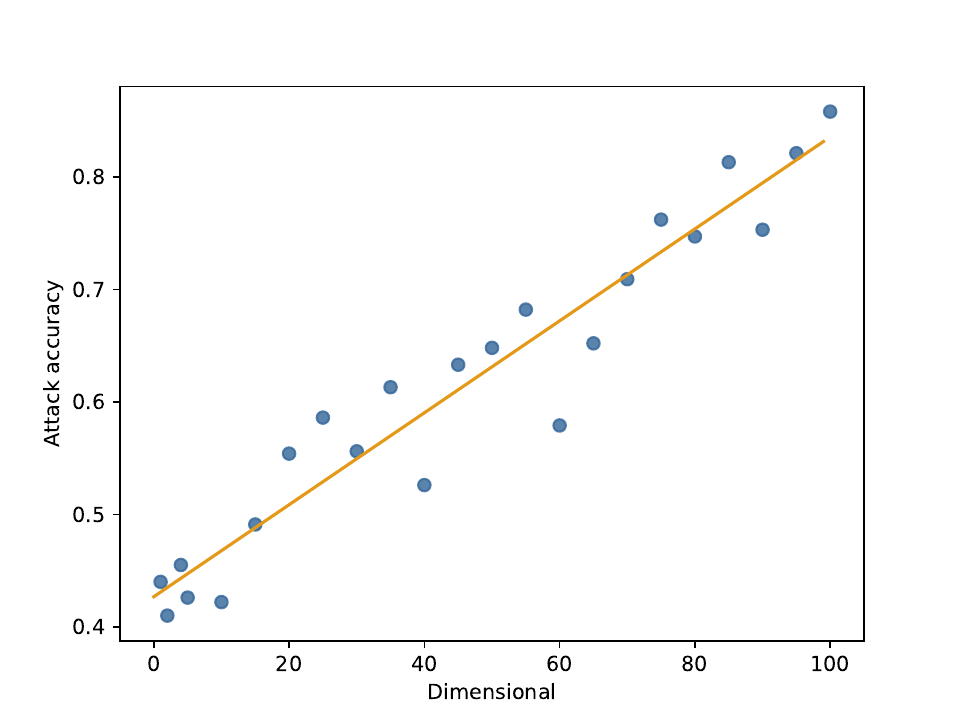}
         \caption{
         The effect of the distribution dimensions on ACC. The ACC is obtained using the target model VGG16 trained on CelebA.}
         \label{fig:dim}
      \end{figure}

\subsubsection{Evaluating the impact of latent distribution dimensions}
This work primarily concerns improving the performance of MI attacks by optimizing the latent distribution in the high-latitude latent space. Therefore, it is necessary to investigate the effect of varying latent distribution dimensions on the MI attack's ACC. Fig.\ref{fig:dim} displays the ACC obtained by DBB-MI, with variations in latent distribution dimensions. As shown in this figure, the ACC of DBB-MI rises gradually with the increase of latent distribution dimensions. This suggests that searching high-dimensional latent distributions can explore the latent space more comprehensively. Hence, MI attacks need to choose the appropriate latent distribution dimension.

\begin{figure}[!t]
    \centering
    \subfloat[Various Structures]{\includegraphics[width=0.47\linewidth]{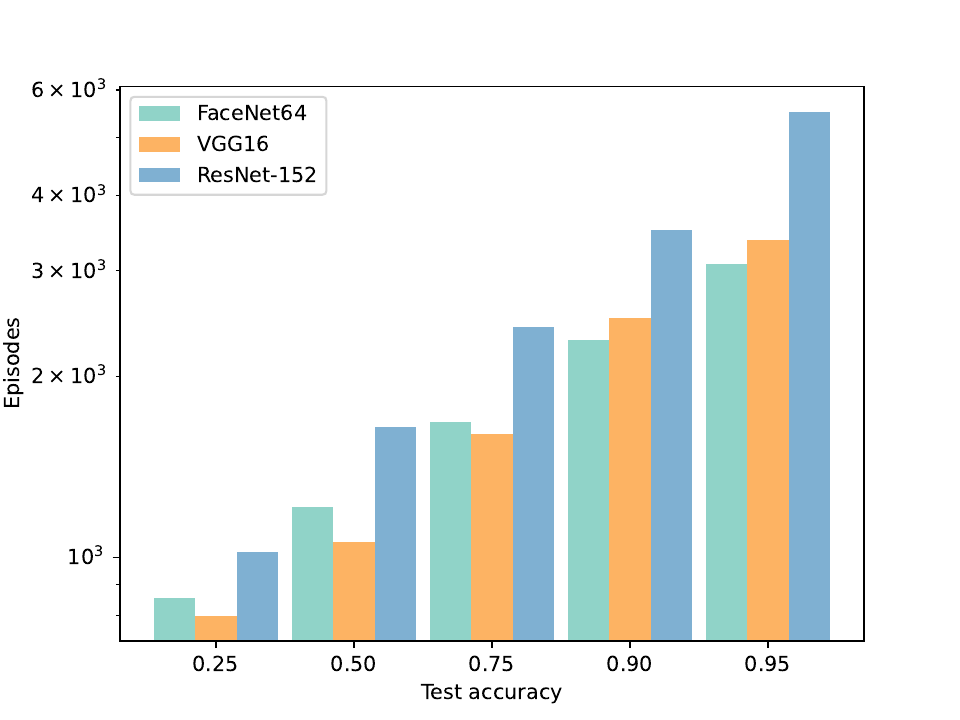}%
    \label{fig:mac}}
    \hfil
    \subfloat[Various Datasets]{\includegraphics[width=0.47\linewidth]{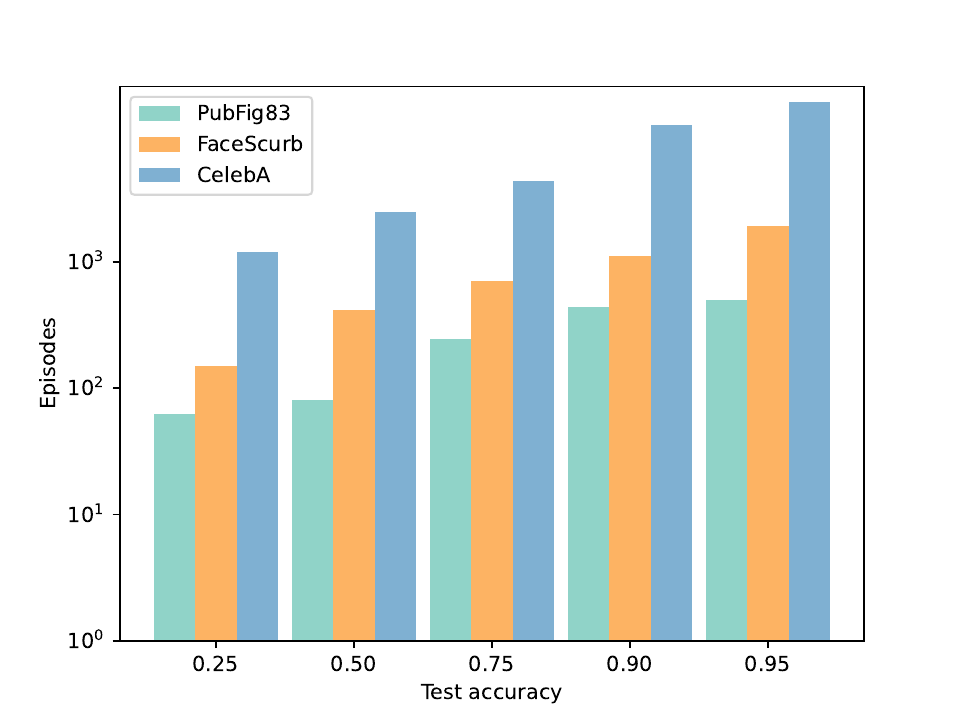}%
    \label{fig:dac}}
     \caption{The median number of iterations required for the generated images to first
     reach a specific test classification accuracy during the attack process. For (a), the dataset is defined as CelebA, and it presents experimental results for different network structures. For (b), the target network is defined as FaceNet64, and the GAN is trained on FFHQ and presents experimental results across various datasets.}
     \label{fig:ac}
  \end{figure}

\subsubsection{Evaluating the impact of training episodes}
To further investigate the factors influencing the performance of MI attacks, we assess the median number of iterations required for generating images to reach the specific test classification model's complexity and the dataset's size. Experimental findings were obtained by training various target models on CelebA. An MI attack has the highest search difficulty to target ResNet-152, the most complex target model. More iterations are required to achieve the test accuracy obtained on FaceNet64 and VGG16, as shown in Fig.\ref{fig:mac}. Fig.\ref{fig:dac} exhibits the experimental results obtained by utilizing a GAN trained on FFHQ to attack FaceNet64 trained on different datasets. An MI attack exhibits the lowest search difficulty and can quickly reach a specific test accuracy on PubFig83, while it needs a higher number of iterations on other datasets. Therefore, we can conclude that as the complexity of the model and the size of the dataset increase, the search difficulty of an MI attack increases, ultimately leading to a decrease in the attack performance.

\begin{figure}[t]
    \centering
    \includegraphics[width=0.8\linewidth]{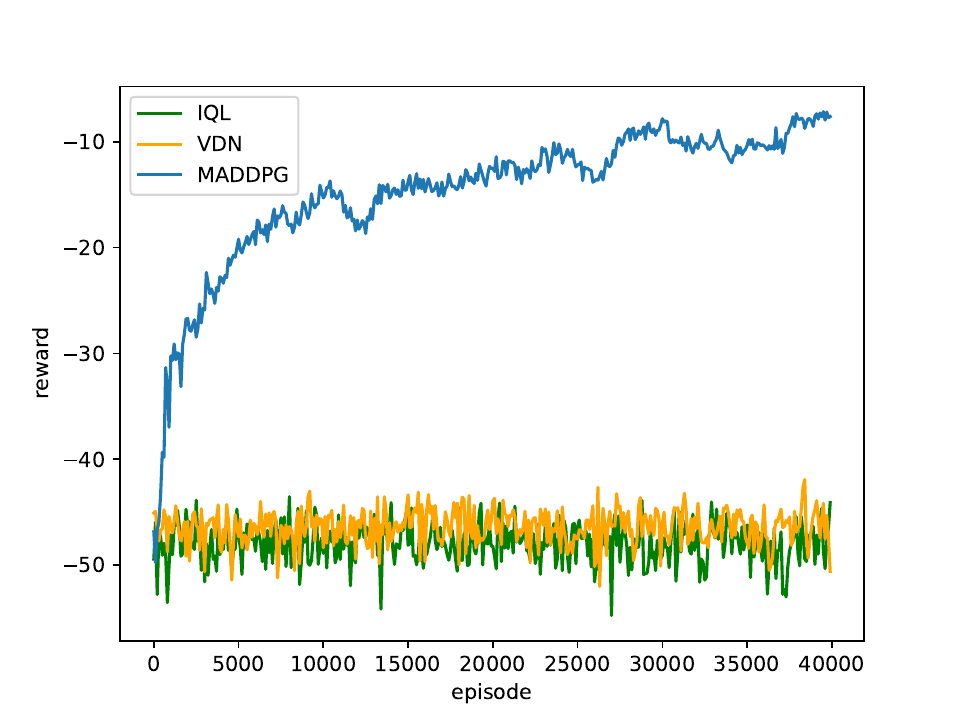}
     \caption{The reward variation of different agents under various episodes. IQL, VDN, and MADDPG represent three different RL agents.}
     \label{fig:reward}
  \end{figure}

\begin{table}[]
\caption{The experimental results of various reinforcement learning agents in MI attacks.}
\small
\centering
    \begin{tabular*}{0.75\linewidth}{cl|cl|cl|cl}
    \toprule
    \multicolumn{2}{c|}{Agent}  & \multicolumn{2}{c}{ACC{$\uparrow$}}   & \multicolumn{2}{c}{PSNR{$\uparrow$}}  & \multicolumn{2}{c}{KNN Dist{$\downarrow$}} \\
    \midrule
    \multicolumn{2}{c|}{IQL}    & \multicolumn{2}{c}{0.341} & \multicolumn{2}{c}{14.21} & \multicolumn{2}{c}{1728.39}  \\
    \multicolumn{2}{c|}{VDN}    & \multicolumn{2}{c}{0.462} & \multicolumn{2}{c}{16.35} & \multicolumn{2}{c}{1443.61}  \\
    \multicolumn{2}{c|}{MADDPG} & \multicolumn{2}{c}{\textbf{0.858}} & \multicolumn{2}{c}{\textbf{20.66}} & \multicolumn{2}{c}{\textbf{1180.63}} \\
    \bottomrule
    \end{tabular*}
    \label{tab:agent} 
\end{table}

  \begin{figure}[t]
    \centering
    \includegraphics[width=0.85\linewidth]{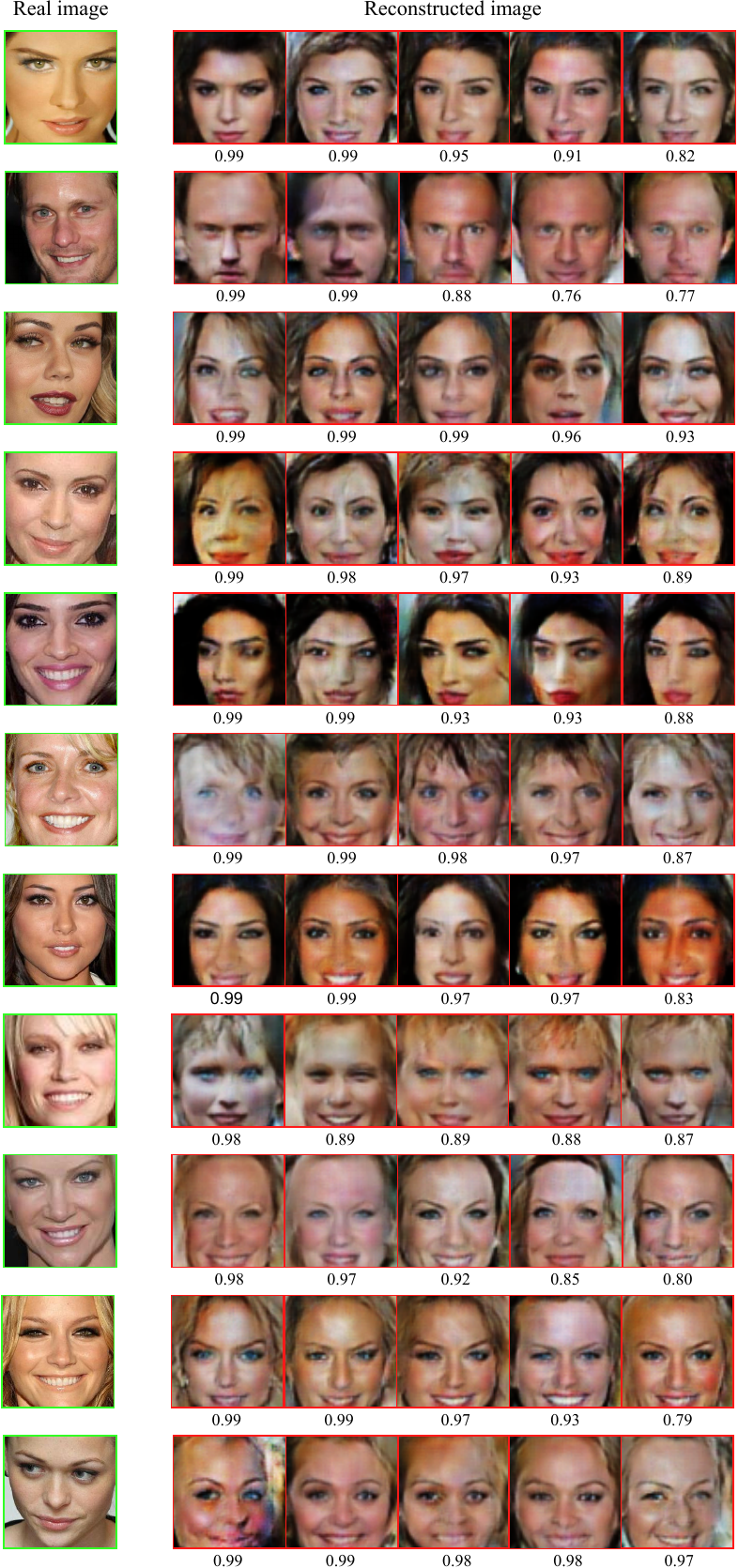}
     \caption{Comparison of original and reconstructed images. In each row, the green box represents the original images with the specific label, and the red box denotes the reconstructed images under the same specific label using DBB-MI. The numbers below the reconstructed images represent the corresponding softmax scores given by the evaluation classifier, indicating that these reconstructed images, to some extent, reveal the privacy information of the specific label.}
     \label{fig:same}
  \end{figure}

\subsubsection{Evaluating various reinforcement learning agents}
DBB-MI heavily relies on the MADDPG to optimize the latent distribution of GAN and obtain more private information about the target. To verify the rationality of using MADDPG, IQL \cite{tampuu2017multiagent},  as a form of fully competitive MARL, and VDN \cite{10.5555/3237383.3238080}, as a form of fully cooperative MARL, are used to search for the latent distribution for model inversion.
Meanwhile, the VGG16 model trained on CelebA is employed as the target network. Table \ref{tab:agent} lists the experimental results of MI attacks using various reinforcement learning agents. This table illustrates that the performance of MADDPG surpasses that of IQL and VDN. For example, MADDPG's ACC is 150\% higher than IQL and 85.7\% higher than VDN, respectively. This is why DBB-MI utilizes MADDPG to optimize GAN's high-dimensional latent space distribution.
Furthermore, this also underscores that searching for suitable latent distributions from the latent space of GANs should be regarded as a semi-competitive, semi-cooperative form of MARL.

\subsubsection{RL agent rewards}
To further assess the performance difference between various agents in GAN-based MI attacks, we compare the reward changes during their training, as shown in Fig.\ref{fig:reward}. As can be seen from the figure, both VDN \cite{10.5555/3237383.3238080}, and IQN \cite{tampuu2017multiagent} exhibit consistently modest rewards, with fluctuations occurring around this baseline amount. In contrast, MADDPG \cite{lowe2017multi} can achieve higher reward convergence, yielding more gratifying results. Therefore, MADDPG is more suitable for hidden space search in GAN-based MI attacks.

\subsubsection{Evaluating the diverse image reconstruction}

It is expected to find several images associated with the same label, depicting various stages or conditions of the same object. DBB-MI can reconstruct several diverse images for a given label, as displayed in Fig.\ref{fig:same}. We sample multiple latent codes from the finally optimized latent distribution to generate multiple images with different privacy attributes. Here, variations in facial expressions, hair, lighting circumstances, and other variables highlight diverse privacy features within the same label. The diverse image reconstruction capabilities of DBB-MI are very important for studying privacy protection. It exposes how an attacker can reconstruct different state information of a target, which also implies how to defend against this attack effectively.

\section{Conclusion}
\label{min:conclusion}
In this paper, we present a novel and effective Distributional Black-Box Model Inversion (DBB-MI) attack that does not require elaborate training of GAN. In a black-box setting, DBB-MI systematically explores the latent space of GAN with limited knowledge to identify the appropriate latent distribution. This is achieved through the utilization of a multi-agent reinforcement learning-based approach. 
It can accurately reconstruct the private data of the target model. 
A comprehensive assessment of the attack performance and generalization of DBB-MI is conducted through a series of experiments. The experimental results demonstrate that DBB-MI attains a level of performance comparable to the most advanced black-box attacks. Additionally, these results further validate the efficacy of distributional attacks in comparison to state-of-the-art MI attacks based on optimizing latent code.




{
    \bibliographystyle{IEEEtran}
    \bibliography{main}
}

\vfill

\end{document}